\def\year{2022}\relax
\documentclass[letterpaper]{article} 
\usepackage{aaai22}  
\usepackage{times}  
\usepackage{helvet}  
\usepackage{courier}  
\usepackage[hyphens]{url}  
\usepackage{graphicx} 
\usepackage{natbib}  
\usepackage{caption} 
\DeclareCaptionStyle{ruled}{labelfont=normalfont,labelsep=colon,strut=off} 
\usepackage{algorithm}
\usepackage{algorithmic}

\usepackage{newfloat}
\usepackage{listings}

\usepackage{xcolor}         
\usepackage{subcaption}
\usepackage{amsmath}
\usepackage{multirow}
\usepackage{url}            
\usepackage{booktabs}       
\usepackage{amsfonts}       
\usepackage{nicefrac}       
\usepackage{microtype}      

\floatstyle{ruled}
\newfloat{listing}{tb}{lst}{}
\floatname{listing}{Listing}
\title{LAMA-Net: Unsupervised Domain Adaptation via Latent Alignment and Manifold Learning for RUL Prediction}
\author {
    Manu Joseph,\textsuperscript{\rm 1}
    Varchita Lalwani, \textsuperscript{\rm 1}
}
\affiliations {
    \textsuperscript{\rm 1} Applied Research, Thoucentric\\
    Bangalore, India\\
    manujoseph@thoucentric.com, varchitalalwani@thoucentric.com
}

\graphicspath{{images/}}
\newcommand{\repo}{\url{https://github.com/repo}}

\begin{document}

\maketitle

\begin{abstract}
  Prognostics and Health Management (PHM) is an emerging field which has received much attention from the manufacturing industry because of the benefits and efficiencies it brings to the table. And Remaining Useful Life (RUL) prediction is at the heart of any PHM system. Most recent data-driven research demand substantial volumes of labelled training data before a performant model can be trained under the supervised learning paradigm. This is where Transfer Learning (TL) and Domain Adaptation (DA) methods step in and make it possible for us to generalize a supervised model to other domains with different data distributions with no labelled data. In this paper, we propose \textit{LAMA-Net}, an encoder-decoder based model (Transformer) with an induced bottleneck, Latent Alignment using Maximum Mean Discrepancy (MMD) and manifold learning is proposed to tackle the problem of Unsupervised Homogeneous Domain Adaptation for RUL prediction. \textit{LAMA-Net} is validated using the C-MAPSS Turbofan Engine dataset by NASA and compared against other state-of-the-art techniques for DA. The results suggest that the proposed method offers a promising approach to perform domain adaptation in RUL prediction. We have made available the code at \repo ~.
\end{abstract}

\section{Introduction}

Prognostics and Health Management (PHM) is an emerging field which has received much attention from the manufacturing industry because of the benefits and efficiencies it brings to the table. The amount of time a machine will probably continue to function before needing repair or replacement is known as the Remaining Useful Life (RUL). RUL prediction is at the heart of any PHM system.

The RUL prediction problem has been addressed in literature with respect to physics, statistical and machine learning methodologies. The failure mechanisms' degradation processes are described mathematically using physics-based methodologies (\citet{physics_based}). These methods demand prior knowledge of deterioration and offer precise RUL estimation when failure can be explained by its physical characteristics. Statistical techniques typically make an effort to fit the data to a probabilistic theory that can capture the uncertainty in the deterioration process (\citet{si2011remaining}). Their flaws are related to data distributions and assumptions regarding changes in health state. In contrast, machine learning models concentrate on discovering the patterns of degradation directly from gathered complex raw data. Machine Learning models like Support Vector Regression (SVR) (\citet{Drucker}), Random Forest (\citet{Breiman2001}), XgBoost (\citet{Chen}), and deep learning models like LSTMs (\citet{Hochreiter1997LongSM}) have been used for prognostic prediction challenges.

Although data-driven approaches based on deep learning have produced promising results for RUL prediction tasks, these approaches require a significant amount of labelled datasets for the network to be trained before it can provide a model that is accurate enough. However, it is frequently challenging to gather enough data with run-to-failure information for complicated systems, especially when companies already use time-based maintenance schedules. Even when we have enough run-to-failure data, models trained on one specific dataset may not generalize well even if some operating conditions change, or some new failure mode is introduced and so on. Although we can start collecting data for the new condition and retrain the models, it usually takes some time to collect the require run-to-failure data. These type of changes (data distribution changes, different input features, limited fault information) is called, collectively, \textit{domain shift}.

In machine learning, one of the ways this shift is tackled is by using \textit{Transfer Learning (TL)}, more specifically \textit{Domain Adaptation (DA)}. DA is a special case of TL where the learning task remains the same, but we move from one domain to another. Several algorithms have been proposed over the years to tackle DA, mostly by reducing domain discrepancy(\citet{gretton2012})(\citet{tzeng2014deep}) between source and target or adversarial learning(\citet{ganin2016domain}). But the large majority of research in the area is in adapting classification tasks with very few addressing regression. And fewer work has targets the specific problem of RUL prediction.

In this work, we propose an unsupervised, homogeneous domain adaptation model with an encoder-decoder structure which uses latent alignment and domain agnostic manifold learning for RUL prediction. Latent alignment is done using Maximum Mean Discrepancy (MMD)(\citet{gretton2012}) and manifold learning is done using a combination of GRU(\citet{cho2014}) based autoencoder and smoothness constraints. We use the C-MAPSS NASA Turbofan degradation dataset (\citet{Saxena2008DamagePM}) to validate our results against other state-of-the-art techniques for DA. We chose this dataset because they contain four run-to-failure data under different operating conditions and failure modes.

The main contributions of the study are:
\begin{itemize}
\item Designing a robust domain-agnostic manifold learning mechanism by the Squeeze Layer and GRU based autoencoder

\item To the best of our knowledge, we are the first to use an autoencoder for domain adaptation for a regression problem.

\item Novel application of smoothness (uniform continuity) constraints for regression tasks
\end{itemize}

The following sections of this article are organized as follows: Section 2 explains the background and related work. Section 3 describes the methodology, and Section 4 explains the experimental results and ablation analysis. Section 5 presents conclusions and future work.

\section{Related Work}

\subsection{RUL Prediction approaches}
The different approaches for RUL prediction can be broadly classified into - model-based (Physics based) (\citet{physics_based}, data-driven (\citet{fang2018review}) and hybrid (\citet{LEI2018799}). Physics-based approaches, although are high fidelity, are usually limited to a few specific components or to similar ones. Data-driven methods are more general because they don't have the strong assumptions Physics-based approaches have.

Statistical methods like survival analysis and hidden markov models have successfully been used in RUL prediction (\citet{si2011remaining}). Modern machine learning approaches like LGBM (\citet{8569801}), XgBoost (\citet{9403743}) and ensembles of Multi-Layered Perceptrons (\citet{5414582}) have been used for RUL Prediction. But the recent surge of interest in deep learning has led to a lot of work in using deep learning models for RUL prediction (\citet{WANG202081}) . \cite{Huang2007} proposed a Feed Forward Network for RUL prediction and showed that it surpassed reliability based approaches. \citet{7998311} proposed a Long Short-Term Memory (LSTM) approach for RUL estimation which made full use of the sensor sequence information and exposed hidden patterns within sensor data with multiple operating conditions, fault and degradation models.(\citet{LI20181}) showed the effectiveness of Deep Convolutional Neural Networks (DCNN) in estimating RUL. (\citet{Peng}) used an LSTM to extract the temporal characteristics of the data sequence, one-dimensional fully-convolutional layer  is adopted to extract spatial features. The spatio-temporal features extracted by the two models are fused and used as the input of the one-dimensional convolutional neural network for RUL prediction. Attention-based DCNN architecture to predict the RUL of turbofan engines with feature ranking metric is explored in (\citet{Muneer}).

Although Transformers(\citet{vaswani2017}) have been used for time series prediction(\citet{autoformer}, \citet{informer}), the focus was mostly on capturing the dependencies between time steps. \citet{dast_zhang} proposed a Transformer based Dual Aspect Self Attention (DAST) model which captured the dependencies between time steps as well as different sensors for RUL prediction.

\subsection{Domain Adaptation for PHM}
All the methods discussed above assumes that sufficient labelled training data is available and that the training and test data are drawn from the same distribution. But in many real-world problems, this is simply not the case. A subset of \textit{Transfer Learning (TL)}, called \textit{Domain Adaptation (DA)} has been developed to address this situation. 

The conventional approach for DA re-weights the source samples by looking at the similarity with target samples(\citet{NIPS2006_a2186aa7}). Subspace Alignment methods (\citet{subspace}) attempt to find a linear map between a number of top eigenvectors by minimizing the Frobenius norm between them. Other approaches attempt to reduce the domain shift problem by minimizing the divergence between source and target features. (\citet{tzeng2014deep}, \citet{Long2015}) used Maximum Mean Discrepancy (MMD) (\citet{gretton2012}) in a CNN architecture with an adaptation layer to learn a representation that is both semantically meaningful and domain invariant on image dataset. CORrelation ALignment or CORAL (\citet{https://doi.org/10.48550/arxiv.1607.01719}) is another simple yet effective method for minimizing the domain shift by aligning the second-order statistics of source and target distributions. Another approach, based on the theory by \citet{Ben-David2010}, is to use a classification loss to directly confuse between domains using adversarial training (\citet{ganin2016domain}, 
\cite{ajakan2014domain}). 

A large majority of research in DA has focused on image datasets and classification problems. Adapting such methods to the RUL prediction task, which has a sequential input instead of a spatial one and is a regression rather than a classification, is not trivial. \citet{Zhang} used a Bi-directional LSTM in a transfer learning paradigm to transfer between domains. But it still needed a few training labels on the target data. It is also worth noting that when transferring from multi-type operating conditions to single operating conditions, transfer learning led to a worse result. \citet{DACOSTA2020106682} has applied unsupervised Domain-Adversarial Neural Networks (DANN) for the RUL prediction domain with some success. (\citet{8942857}) focuses on a domain adaptation RUL prediction model by integrating the adaptive batch normalization (AdaBN) (\citet{LI2018109}) into DCNN. \citet{9419848} proposed Deep Subdomain Adaptive Regression Network(DSARN) which tries to make the alignment between source and target on sub-domains rather than global alignment. \citet{9234721} proposed a two-step domain adaptation technique using adversarial domain adaptation and contrastive loss to generalize between target and source.

\subsection{Domain Adaptation through Manifold Learning}
Autoencoders have been used for manifold learning (\citet{wei2016locally}) in the semi-supervised setting. They have also been used for domain adaptation (\citet{Deng2014AutoencoderbasedUD}. The success of such approaches lies in the ability of the autoencoder to learn the underlying manifold in the input space. Many approaches also uses explicit manifold geometry based alignment functions (\citet{Baktash2014DomainAO}, \citet{Wang2018VisualDA}) to learn a domain agnostic manifold and perform domain adaptation through it. But these approaches usually introduce some computational instability and complexity.

\section{LAMA-Net: Unsupervised Domain Adaptation via Latent Alignment and Manifold Learning}

In this section, we present \textit{LAMA-Net} (Unsupervised Domain Adaptation via \textbf{L}atent \textbf{A}lignment and \textbf{MA}nifold Learning), our domain adaptation model to predict the RUL of assets across different domains with different fault modes and operating conditions. We will first introduce, formally, the problem definition and then go on in detail about the model architecture, the different components used, and the training and optimization.

\subsection{Problem Definition}
Given a source domain, $\mathcal{D}_s$, and a corresponding learning task, $\mathcal{T}_s$, a target domain, $\mathcal{D}_t$ and a learning task, $\mathcal{T}_t$. For the domain adaptation setting, we have $\mathcal{T}_s=\mathcal{T}_t$ and $\mathcal{D}_s \neq \mathcal{D}_t$.

We make the following assumptions/notations:
\begin{itemize}
    \item The source domain, $\mathcal{D}_s = \left\{(x^i_s, y^i_s)\right\}_{i=1}^{N_s}$, contain $N_s$ training examples and $x_s^i \in \mathcal{X}_s$ is a sequence of multivariate time-series sensor data of length $K_t$ and $f$ features, i.e. $x_s^i \in \mathbb{R}^{f \times K_t}$. $y^i_s \in \mathcal{Y}_s$ denotes the Remaining Useful Life (RUL) for each sequence $x_s^i$
    \item The target domain, $\mathcal{D}_t = \left\{(x^i_t)\right\}_{i=1}^{N_t}$, contain $N_t$ training examples and $x_t^i \in \mathcal{X}_t$ is a sequence like the target domain but without a target value (\textit{unsupervised domain adaptation})
    \item $\mathcal{D}_s$ and $\mathcal{D}_t$ are sampled from distinct marginal probability distributions $\mathbb{P}(\mathcal{X}_s) \neq \mathbb{P}(\mathcal{X}_t)$
    \item $x^i_t \in \mathbb{R}^{f \times K_t}$ to have the same number of features as the source domain (\textit{homogeneous domain adaptation}), but sampled from a different probability distribution
    \item We denote the number of training examples for source and target separately as $N_s$ and $N_t$ purely for the ease of exposition. Practically, while training, $N_s = N_t$.
\end{itemize}

Our goal is to learn a function, $g$, which approximates the RUL in the target domain at testing time directly from degradation data, i.e. $y^i_t \approx g(x^i_t)$. We have considered the length of source sequence, $K_s$, and target sequence, $K_t$, to be same but it can be different as per domain and the proposed method would be able to handle such differences using padding.

\subsection{Model Architecture}

\begin{figure*}[!ht]
    \centering
    \includegraphics[width=0.7\linewidth]{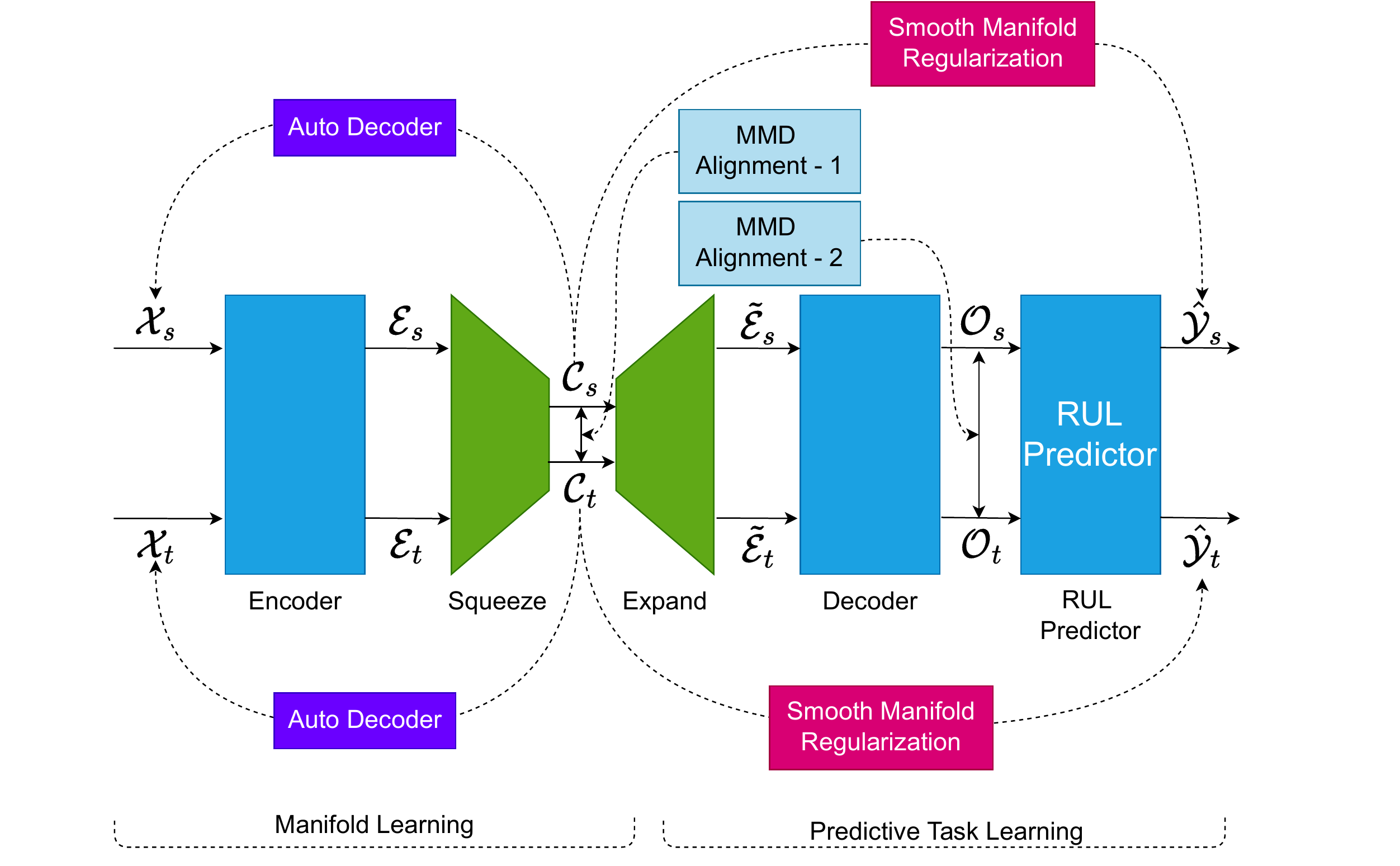}
    \caption{LAMA-Net Architecture}
    \label{fig:model_arch}
\end{figure*}

The model architecture (Figure \ref{fig:model_arch}) follows a Siamese-twins like parallel streams which processes source and target domain input examples. All the components in the network architecture is shared between the source and target domains with sufficient incentives for the latent representations to be domain agnostic so that we can generalize across domains. The model architecture has these main components:
\begin{itemize}
    \item Encoder
    \item Squeeze and Expand Layers
    \item Decoder
    \item Latent Alignment via Maximum Mean Discrepancy Loss
    \item Reconstruction Loss
    \item Manifold Learning
    \item Loss Function
\end{itemize}

\subsection{Encoder}

The model architecture uses an encoder-decoder framework to encode the raw input signals into a latent space and then decode the latent representation into RUL prediction. We represent the \textit{encoder} as a non-linear function, $\mathit{Enc}$, which transforms the input, $\mathcal{X} \in \mathbb{R}^{f \times K}$ to a latent representation, $\mathcal{E} \in \mathbb{R}^M$, where $M$ is the latent dimension. Formally, for both the source and target domain, we have

\begin{align}
    &\mathcal{E}_s = \mathit{Enc}(\mathcal{X}_s) &&
    \mathcal{E}_t = \mathit{Enc}(\mathcal{X}_t)
\end{align}

Although \textit{LAMA-Net} is agnostic to the kind of encoder-decoder we are using, for the purposes of this paper we have chosen a transformer based encoder-decoder model which has showed state-of-the-art performance in RUL prediction - Dual Aspect Self-Attention based on Transformer (DAST) for Remaining Useful Life Prediction(\citet{dast_zhang}). DAST model is a transformer-based model which uses self attention that captures the weighted features from both the sensor and time step dimensions. The DAST encoder consists of a sensor encoder layer, time step encoder layer, and a feature fusion layer, which combines the two generated weighted features from sensor and time step encoder layers. The DAST decoder is a similar to the original Transformer(\citet{vaswani2017}).

\subsection{Squeeze and Expand Layer}

Depending on the type of encoder function we use, $M$, the latent dimension, can be arbitrarily large of small. To be truly agnostic of the type of encoder used, we have introduced explicit \textit{Squeeze} and \textit{Expand} Layers to introduce a bottleneck in the information flow. This bottle neck is essential so that the alignment we want across the source and target domains can be done in a space without any noise and be more efficient because of it.
The \textit{Squeeze} and \textit{Expand} Layers are layered feed forward networks of the form, $\sigma(\mathbb{W}x+b)$, where $\sigma$ is any non-linear activation function. 

Let us represent the non-linear function learned by the \textit{Squeeze Layer} as $\mathit{S}$ and the \textit{Expand Layer} as $\mathit{E}$. The Squeeze Layer takes in the latent representation from the encoder, $\mathcal{E}$, and squeezes representation into $\mathcal{C} \in \mathbb{R}^B$, where $B$ is the bottleneck dimension and $B<<M$. Formally, for both source and target domains, we have


\begin{align}
    &\mathcal{C}_s = \mathit{S}(\mathcal{E}_s) &&
    \mathcal{C}_t = \mathit{S}(\mathcal{E}_t)
\end{align}
The \textit{Expand Layer} ($\mathit{E}$) takes in the bottleneck representation, $\mathcal{C} \in \mathbb{R}^B$, and transforms it back to the same dimension as the initial latent representation, $\tilde{\mathcal{E}} \in \mathbb{R}^M$.Formally, for both source and target domains, we have

\begin{align}
    &\tilde{\mathcal{E}}_s = \mathit{E}(\mathcal{C}_s) &&
    \tilde{\mathcal{E}}_t = \mathit{E}(\mathcal{C}_t)
\end{align}
\subsection{Decoder and RUL Prediction}

The decoder($\mathit{D}$) takes the expanded latent representation, $\tilde{\mathcal{E}} \in \mathbb{R}^M$, and transforms into a representation suitable for the linear RUL prediction layer ($\mathit{R}$), $\mathcal{O} \in \mathbb{R}^P$, where $P$ is the dimension of the hidden representation just before the linear RUL prediction layer. Formally, for both source and target domains, we have
\begin{align}
    &\mathcal{O}_s = \mathit{D}(\tilde{\mathcal{E}}_s) &&
    \mathcal{O}_t = \mathit{D}(\tilde{\mathcal{E}}_t)
\end{align}
The final linear RUL Prediction layer takes in $\mathcal{O} \in \mathbb{R}^P$ and transforms it into the final RUL Prediction, $\mathcal{Y} \in \mathbb{R}^1$, using a simple linear layer with a weight matrix, $W_R$, a bias matrix, $b_R$, and a non-linear activation function, $\sigma$. Formally, for both source and target domains, we have

\begin{equation}
\begin{split}
    \hat{\mathcal{Y}}_s = \mathit{R}(\mathcal{O}_s) = \sigma(W_R \mathcal{O}_s + b_R) \\
    \hat{\mathcal{Y}}_t = \mathit{R}(\mathcal{O}_t) = \sigma(W_R \mathcal{O}_t + b_R)
\end{split}
\end{equation}
The $\hat{\mathcal{Y}}_s$ is compared to the actual RUL, $\mathcal{Y}_s$ using the standard mean squared error (MSE). 
\begin{equation}
    \mathcal{L}_{RUL} = \frac{1}{N_s} \sum_{i=1}^{N_s} \left(y_s^i - \hat{y}_s^i\right)^2
\end{equation}

This loss ($\mathcal{L}_{RUL}$) ensures the network learns the right kind of representations so that the final task of predicting the RUL is learned well enough.

\subsection{Latent Alignment via Maximum Mean Discrepancy}
We know that $\mathcal{D}_s$ and $\mathcal{D}_t$ are sampled from distinct marginal probability distributions, $P(\mathcal{X}_s)$ and $P(\mathcal{X}_t)$. Because of the weight sharing in the proposed network generalization from $\mathcal{D}_s$ to $\mathcal{D}_t$ can happen efficiently only when the network learns domain agnostic latent representations. And to this effect, our proposed method strives to align the probability distributions of latent representations, namely $\mathcal{C}$ and $\mathcal{O}$, using Maximum Mean Discrepancy (MMD). 

Maximum Mean Discrepancy (MMD) is an effective non-parametric metric for measuring the discrepancy between two marginal probability distributions by embedding distributions in a reproducible kernel Hilbert space (RKHS) (\citet{gretton2012}). Kernel embeddings of distributions allows us to embed a distribution, $\mathbb{P}$, into an RKHS, $\mathcal{H}_{\phi}$, with a kernel $\phi$ (\citet{sriperumbudur2010}). In the RKHS, $\mathbb{P}$ is represented as an element, $\mu_{\mathbb{P}}$. Formally, $\mu_{\mathbb{P}} := \mathbb{E}_{x \sim \mathbb{P}}\left[\phi(.,x) \right] \in \mathcal{H}_{\phi}$, where $x \sim \mathbb{P}$ means that random variable $x$ has probability distribution $\mathbb{P}$, $\mathbb{E}_{x \sim \mathbb{P}}\left[f(x) \right]$ denotes the expectation of function f with respect to the random variable $x$. The kernel embedding captures all the properties about the distribution like the mean, variance, and other higher order moments if $\phi$ is \textit{characteristic} (e.g. RBF Kernel) (\citet{sriperumbudur2010}. Now we can define MMD as the RKHS norm between the kernel embeddings of two distributions.

For the source and target domains, we can formally define MMD as,
\begin{equation}
\begin{split}
    MMD(\mathbb{P}_s, \mathbb{P}_t) = \Vert \mu_{\mathbb{P}_s} -\mu_{\mathbb{P}_t}\Vert_{\mathcal{H}_{\phi}} \\
    = \Vert \mathbb{E}_{x \sim \mathbb{P}_s} \left[ \phi(.,x)\right] - \mathbb{E}_{y \sim \mathbb{P}_t} \left[ \phi(.,y)\right] \Vert_{\mathcal{H}_{\phi}}
\end{split}
\end{equation}
We have $MMD(\mathbb{P}_s, \mathbb{P}_t) =0$ \textit{iff} $\mathbb{P}_s=\mathbb{P}_t$. If we denote $\mathcal{C}_s = \left\{x^i_s\right\}_{i=1}^{N_s}$ and $\mathcal{C}_t = \left\{x^i_t\right\}_{i=1}^{N_t}$ as two sets of samples drawn \textit{i.i.d} from distributions \textit{s} and \textit{t} respectively, the empirical estimate of MMD can be given by (\citet{gretton2012}),
\begin{equation}
    MMD^2(\mathcal{C}_s, \mathcal{C}_t) = \left\Vert \frac{1}{N_s} \sum^{N_s}_{i=1} \phi(x_s^i) -  \frac{1}{N_t} \sum^{N_t}_{i=1} \phi(x_t^i) \right\Vert^2_{\mathcal{H}_\phi}
\label{mmd_eqn}
\end{equation}

We can cast Eq.\ref{mmd_eqn} into a vector-matrix multiplication form and come up with the \textit{kernelized} equation (\citet{borgwardt2006})
\begin{equation}
\begin{split}
    MMD^2(\mathcal{C}_s, \mathcal{C}_t) = \left( \frac{1}{N_s^2} \sum^{N_s}_{i=1}\sum^{N_s}_{j=1} \phi(x_s^i,x_s^j) + \right.\\ 
    \left.\frac{1}{N_t^2} \sum^{N_t}_{i=1}\sum^{N_t}_{j=1} \phi(x_t^i,x_t^j) -  \frac{2}{N_t N_s} \sum^{N_s}_{i=1}\sum^{N_t}_{j=1} \phi(x_s^i,x_s^j) \vphantom{\frac{1}{N_s^2}} \right)
\end{split}
\end{equation}

We use MMD to align the feature distribution of two latent representations, $\mathcal{C}_s$ and $\mathcal{C}_t$ as well as $\mathcal{O}_s$ and $\mathcal{O}_t$ to make the learned latent representations domain agnostic. Formally,
\begin{equation}
    \mathcal{L}_{MMD} = MMD(\mathcal{C}_s, \mathcal{C}_t) + MMD(\mathcal{O}_s, \mathcal{O}_t)
\end{equation}

\subsection{Reconstruction Loss}
To induce more structure into the compressed latent representation, $\mathcal{C}$, we propose to add a reconstruction loss ($\mathcal{L}_{recon}$) which uses a standard GRU(\citet{cho2014}) based decoder to reconstruct the input, $\mathcal{X}$. We construct the initial hidden state of the GRU from the compressed latent, $\mathcal{C}$ using a simple linear layer. Formally, we can represent this as
\begin{align}
    &h_s = \sigma (W_r \mathcal{C}_s + b_r) &&
    h_t = \sigma (W_r \mathcal{C}_t + b_r)
\end{align}
Without going into details of the gating mechanism, the learned function by the GRU can be represented as,
\begin{align}
    &\hat{\mathcal{X}}_s = g(h_s; \theta_r) &&
    \hat{\mathcal{X}}_t = g(h_t; \theta_r)
\end{align}
We use the first element in the sequence and the hidden state constructed from $\mathcal{C}$ to reconstruct the input sequence.The reconstruction loss ($\mathcal{L}_{recon}$) is the standard MSE loss and is represented formally as
\begin{equation}
    \mathcal{L}_{recon} = \frac{1}{N_s} \sum_{i=1}^{N_s} \left(x_s^i - \hat{x}_s^i\right)^2 +  \frac{1}{N_t} \sum_{i=1}^{N_t} \left(x_t^i - \hat{x}_t^i\right)^2
\end{equation}
\subsection{Manifold Learning and Smoothness Constraint}
The manifold hypothesis states that many high-dimensional datasets from the real world, actually lie along a low-dimensional manifold embedding inside that high-dimensional space (\citet{Cayton05algorithmsfor}). We hypothesize that the encoder ($\mathit{Enc}$) and the \textit{Squeeze Layer} ($\mathit{S}$), along with the GRU-based auto-encoder discovers the embedded manifold ($\mathcal{M}$) in the input space ($\mathcal{X}$) and learns a mapping to Euclidean space, $\mathbb{R}^M$. The latent alignment via MMD encourages the network to learn a domain-agnostic manifold which helps us generalize to new domains. 

To add further structure to the learned manifold projection ($\mathcal{C}$), we impose another constraint such that the mapping $\mathcal{C} \longrightarrow \mathcal{Y}$ is uniformly continuous\footnote{We thank Jan Jitse Venselaar for his help with a bit of math here}, i.e. if $x \in \mathcal{C} \longrightarrow y \in \mathcal{Y}$ then points in the immediate neighbourhood of $x$ should map to points which are in the immediate neighbourhood of $y$. Intuitively, we can think of this as a "smoothness" constraint (although it need not be smooth in the mathematical sense). We do this by introducing another loss,
\begin{equation}
    \mathcal{L}_{smooth} = \left \Vert \mathit{F}(\mathcal{C}) - \mathit{F}(\mathcal{C} + \gamma_{noise} \delta) \right \Vert^2
\end{equation}

where, $\mathit{F}(.) = \mathit{R}(\mathit{D}(\mathit{E}(.)))$, $\delta \in \mathcal{N}(0,1)$ is a Gaussian noise vector we use to perturb $\mathcal{C}$, and $\gamma_{noise}$ is the scaling factor with which we decide the locality of the perturbation.

\begin{figure}[!ht]
    \centering
    \includegraphics[width=1\linewidth]{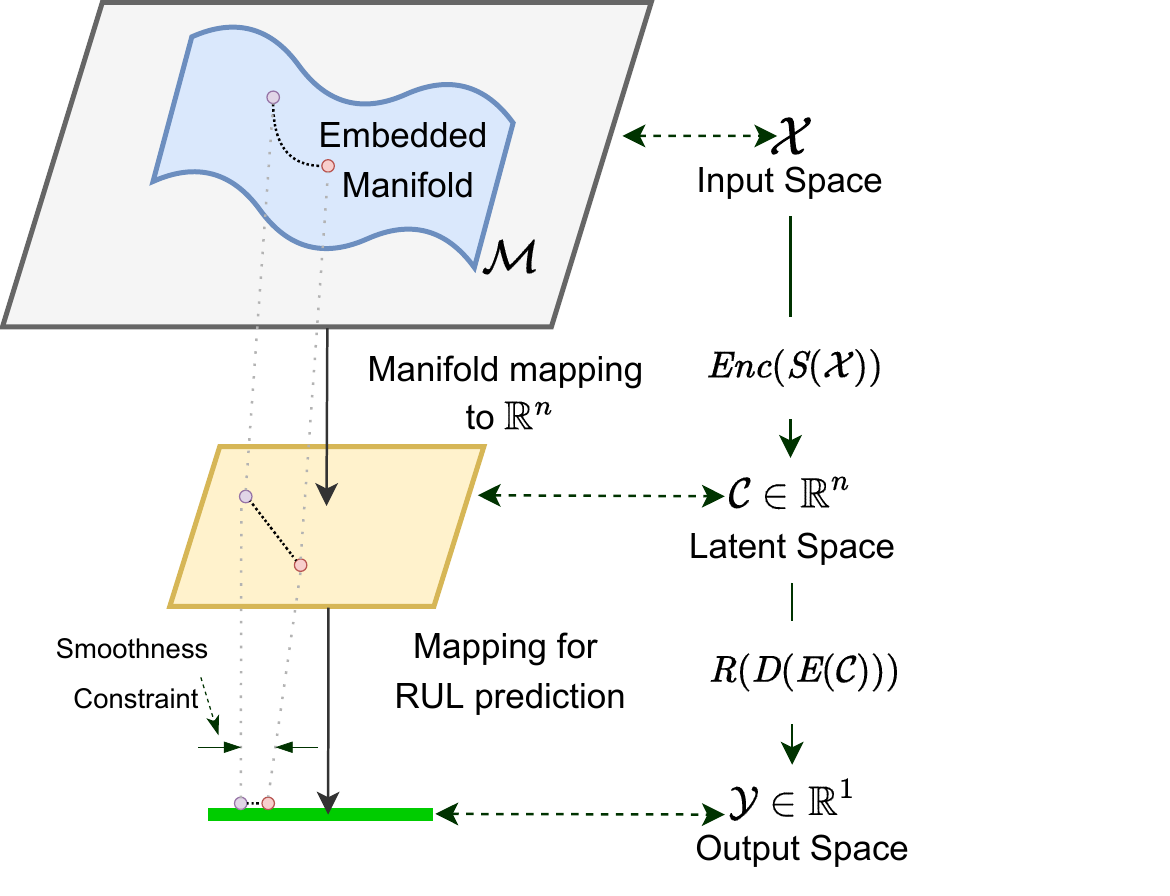}
    \caption{LAMA-Net Architecture: Manifold Perspective}
    \label{fig:model_arch_manifold}
\end{figure}

\subsection{Loss and Optimization}

The final loss function of the \textit{\textit{LAMA-Net}} is:
\begin{equation}
\begin{split}
    \mathcal{L} = \mathcal{L}_{RUL}^S + \lambda_{m} \mathcal{L}_{MMD} + \lambda_{r} \mathcal{L}_{recon}^S +\\ 
      \lambda_{r} \mathcal{L}_{recon}^T + \lambda_{s} \mathcal{L}_{smooth}^T + \lambda_{s} \mathcal{L}_{smooth}^T
\end{split}
\end{equation}
where, $\lambda_{m},\text{ } \lambda{r},\text{ and } \lambda{s}$ are hyperparameters with which we can control how important each term in the loss function is. The domain adaptation losses only kick in after the first 200 iterations of training.

Since our architecture is end-to-end differentiable, we can easily optimize the loss, $\mathcal{L}$, using gradient-based methods like gradient descent.

\section{Experiments}
 In this section, we describe the data set used for the experiments, the experiment procedure and details about the implementation. We also report the comparison of \textit{LAMA-Net} with other state-of-the-art DA techniques. We also provide ablation studies and other investigation to demonstrate the importance of the architecture design decisions.

\subsection{C-MAPPS Datasets}
To evaluate \textit{LAMA-Net}, we have selected the popular C-MAPSS(\citet{Saxena2008DamagePM}) benchmark dataset of run-to-failure data of turbofan engines, simulated using a thermo-dynamic simulation model. The dataset contains information from 21 sensors as well as 3 operating conditions and is composed of four subsets, namely FD001, FD002, FD003 and FD004, which differ in terms of working conditions, fault modes, life spans, and number of engines (Table \ref{tab: datasets}). The operating conditions vary from one in FD001 and FD003, to six in FD002 and FD004. For our experiments, we consider each one of the subsets as source and target domains and perform domain adaptation on all different source-target pairs. Appendix has details about pre-processing.

\subsection{Experimental Study and Analysis}
All experiments used PyTorch (\citet{NEURIPS2019_9015}) and were run on a single NVIDIA K80/T4 GPU with 2 cores and 12GB of RAM. The batch size was fixed at \textit{128}. Initial learning rate was fixed as 1e-3 with a standard Adam optimizer. Exponential Decay with $\gamma = 0.95$, which started the decay after 100 iterations, was used as the learning rate decay for all experiments. We use RMSE and and an asymmetric scoring function (\citet{Saxena2008DamagePM}) as the two metrics to evaluate different models. Details regarding the performance metrics, training procedure, hyperparameter selection and final hyperparameters are present in the Appendix.

\subsection{Comparison With State-of-the-Art Methods}
To evaluate and benchmark \textit{LAMA-Net}, we have chosen to keep the same structure so that the learning capacity of the model remains the same, but compare only the domain adaptation techniques. The autoencoder and the smoothing constraint in \textit{LAMA-Net} are the key aspects of our setup. Therefore, we compare the performance of \textit{LAMA-Net} with other alternatives like DANN (\citet{ganin2016domain}), MMD (\citet{gretton2012})(\citet{tzeng2014deep}), CORAL (\citet{https://doi.org/10.48550/arxiv.1607.01719}), and without domain adaptation (No DA) using same DAST architecture with bottlenecks. Table \ref{tab: rmse_ranking} and \ref{tab: score_ranking} shows the RMSE and Score for all the source-target combinations using all these modelling alternatives. The standard deviation of Score across the three random seeds are also included in the appendix. Key observations from the results are:
\begin{itemize}
    \item We can see that \textit{LAMA-Net} is clearly superior to alternatives. Both in RMSE and Score, \textit{LAMA-Net} performs better than the alternatives in a large majority of source-target combinations. And even in cases where \textit{LAMA-Net} is performing slightly worse than an alternative, it makes up for that with increased resilience to random seeds.
    \item Comparing using the score function makes the superiority even more clearer as the scores of \textit{LAMA-Net} is orders of magnitude better than alternatives, making it the safer choice in any real implementation.
    \item Even between difficult pairs (FD003-FD002, FD004-FD001, etc) \textit{LAMA-Net} consistently gives better adaptation results. We hypothesize that FD003-FD001 is an easy adaptation task and therefore all the adaptation techniques fail to beat the model which has no adaptation.
\end{itemize}

\begin{table*}[!ht]
\centering
\caption{Average RMSE with S.D for RUL Prediction across all source-target combinations (3 Random seeds)}
\label{tab: rmse_ranking}
\centering
\begin{tabular}{llccccc}
	\toprule[1.5pt]
	\rule{1pt}{0pt}{Source} & Target & No DA            & DANN             & MMD              & CORAL            & LAMA-Net            \\
	\midrule[1pt]
	FD001                   & FD002  & 96.04$\pm$1.96   & 97.92$\pm$25.65  & 100.31$\pm$11.2  & 105.11$\pm$10.42 & \textbf{53.88$\pm$0.01}  \\
	FD001                   & FD003  & 50.04$\pm$3.7    & 44.62$\pm$4.5    & 45.57$\pm$8.73   & \textbf{35.76$\pm$8.88}   & 40.54$\pm$1.24  \\
	FD001                   & FD004  & 110.38$\pm$4.93  & 101.51$\pm$19.18 & \textbf{51.43$\pm$45.22}  & 94.28$\pm$9.27   & 55.26$\pm$0.08  \\
	FD002                   & FD001  & 103.26$\pm$19.79 & 128.13$\pm$27.55 & 118.25$\pm$17.64 & 122.43$\pm$37.47 & \textbf{41.87$\pm$0.21 } \\
	FD002                   & FD003  & 104.79$\pm$8.39  & 95.68$\pm$16.99  & 112.11$\pm$11.93 & 123.04$\pm$41.84 & \textbf{41.9$\pm$0.4}    \\
	FD002                   & FD004  & 75.45$\pm$25.7   & 75.9$\pm$24.42   & 75.3$\pm$25.44   & 74.98$\pm$18.74  & \textbf{35.61$\pm$7.53}  \\
	FD003                   & FD001  & \textbf{27.9$\pm$5.53}    & 63.27$\pm$4.99   & 47.03$\pm$10.15  & 70.25$\pm$54.09  & 41.01$\pm$11.89 \\
	FD003                   & FD002  & 103.12$\pm$9.98  & 109.56$\pm$13.96 & 104$\pm$16.34    & 100.3$\pm$31.2   & \textbf{57.05$\pm$3.75}  \\
	FD003                   & FD004  & 78.81$\pm$35.47  & 84.02$\pm$26.42  & 99.73$\pm$14.99  & 154.39$\pm$52.68 & \textbf{50.89$\pm$8.52}  \\
	FD004                   & FD001  & 116.19$\pm$36.27 & 165.96$\pm$18.02 & 181.52$\pm$9.2   & 118.59$\pm$22.01 & \textbf{42.78$\pm$0.37}  \\
	FD004                   & FD002  & 46.32$\pm$16.01  & 60.79$\pm$8.81   & 80.47$\pm$6.41   & 83.46$\pm$7.86   & \textbf{34.8$\pm$0.37}   \\
	FD004                   & FD003  & 100.35$\pm$24.6  & 129.71$\pm$59.84 & 153.38$\pm$23.24 & 108.62$\pm$56.18 & \textbf{42.76$\pm$0.52}  \\
	\bottomrule[1.5pt]
\end{tabular}
\end{table*}

\begin{table*}[!ht]
\centering
\caption{Average Score for RUL Prediction across all source-target combinations (3 Random seeds)}
\label{tab: score_ranking}
\centering
\begin{tabular}{llccccc}
	\toprule[1.5pt]
	\rule{1pt}{0pt}{Source} & Target & No DA                 & DANN                  & MMD                   & CORAL                 & LAMA-Net                  \\
	\midrule[1pt]
	FD001                   & FD002  & 9.25E+18 & 1.65E+11 & 8.56E+11 & 1.48E+17 & \textbf{9.23E+05} \\
	FD001                   & FD003  & 2.83E+05 & 1.35E+06 & 3.77E+05 & 1.11E+06 & \textbf{2.24E+0}4 \\
	FD001                   & FD004  & 5.61E+21& 3.64E+10 & 4.97E+10 & 4.16E+14 & \textbf{9.99E+05} \\
	FD002                   & FD001  & 9.72E+09& 3.41E+17 & 1.15E+16 & 1.48E+16 & \textbf{2.07E+04} \\
	FD002                   & FD003  & 2.31E+08& 1.16E+13 & 1.89E+16 & 2.29E+18 & \textbf{2.30E+04} \\
	FD002                   & FD004  & 6.74E+28& 2.69E+27 & 5.13E+27 & 3.12E+26 & \textbf{1.51E+0}5 \\
	FD003                   & FD001  & \textbf{1.33E+04}& 2.04E+10 & 8.31E+07 & 1.46E+09 & 1.51E+05 \\
	FD003                   & FD002  & 7.87E+18& 5.06E+10 & 2.46E+10 & 8.52E+17 & \textbf{8.62E+06} \\
	FD003                   & FD004  & 8.06E+19& 3.32E+10 & 9.27E+09 & 1.93E+17 & \textbf{5.90E+06} \\
	FD004                   & FD001  & 1.12E+10& 7.57E+10 & 6.76E+11 & 3.31E+14 & \textbf{3.17E+04} \\
	FD004                   & FD002  & 4.21E+23& 2.55E+21 & 2.17E+25 & 1.23E+30 & \textbf{6.14E+04} \\
	FD004                   & FD003  & 3.61E+09& 2.38E+11 & 8.31E+11 & 6.84E+12 & \textbf{3.14E+04} \\
	\bottomrule[1.5pt]
\end{tabular}
\end{table*}

\subsection{Ablation Analysis}
In this section, we analyze the contribution and effect of major architectural decisions, such as adding an autoencoder and the smoothness constraint. We are comparing three versions of our proposed model:

\begin{itemize}
    \item \textbf{MMD}: This version just has the MMD based latent alignment
    \item \textbf{MMD + Autoencoder}: This version has the GRU based autoencoder along with the MMD latent Alignment
    \item \textbf{MMD + Autoencoder + Smoothness}: This is the final version, which is what we call \textit{LAMA-Net}.
\end{itemize}

\begin{figure}[!ht]
    \centering
    \includegraphics[width=1\linewidth]{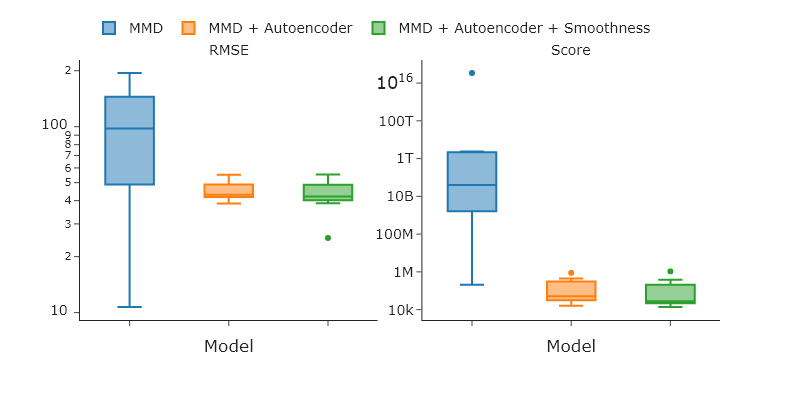}
    \caption{RMSE and Score of different versions of the proposed model}
    \label{fig:ablation_rmse_score}
\end{figure}

The RMSE and Score distribution across all source-target combinations is depicted in Figure \ref{fig:ablation_rmse_score}. We can see that adding the Autoencoder  brings with it the maximum benefit as there is a large shift in means and reduction in variation between \textbf{MMD} and \textbf{MMD + Autoencoder}, in both RMSE and Score.The addition of the smoothness constraint also brings, but notably smaller, improvement in RMSE and Score.

\begin{figure}[!ht]
    \centering
    \includegraphics[width=1\linewidth]{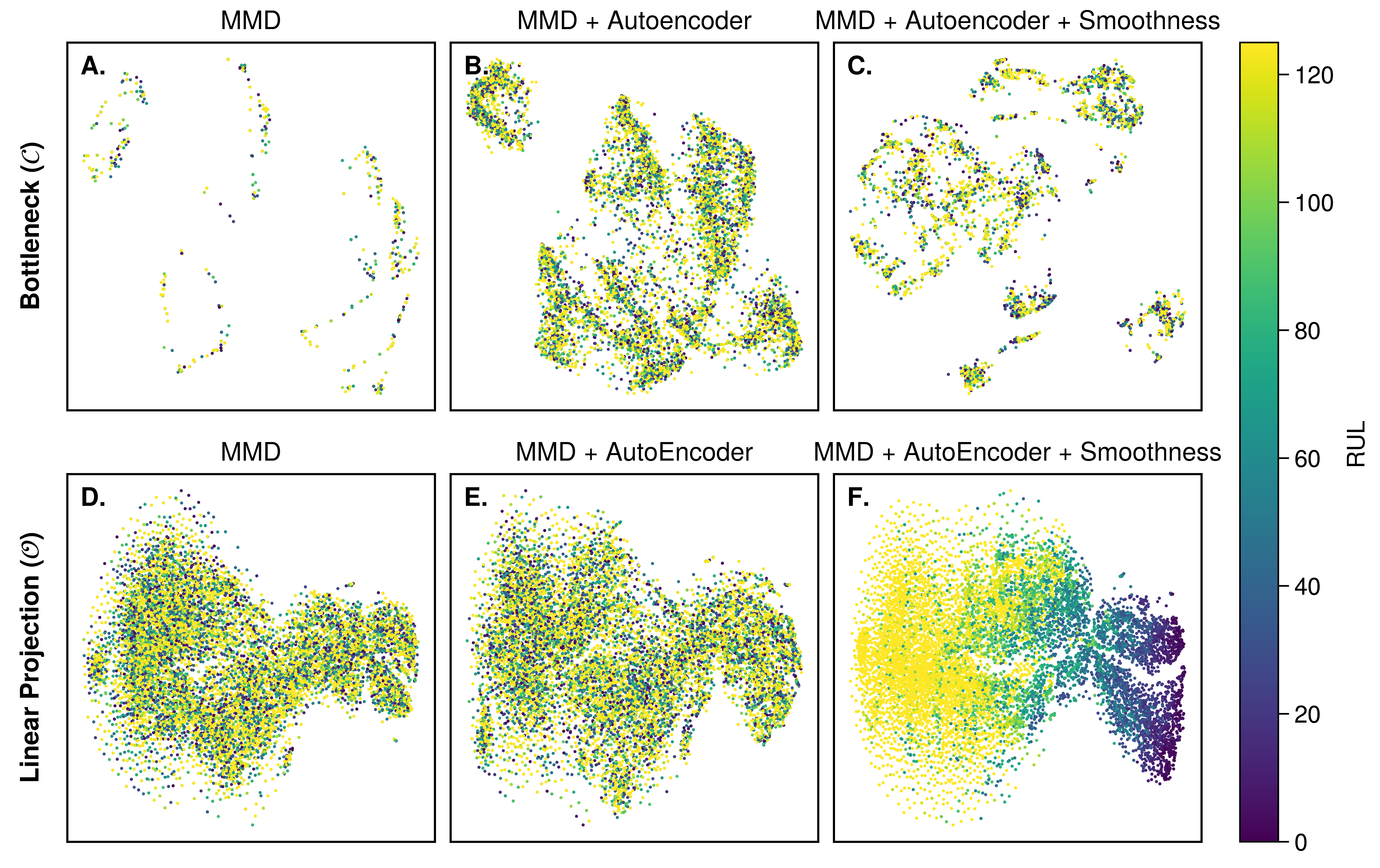}
    \caption{t-SNE Visualization of $\mathcal{C}$ and $\mathcal{O}$ for different versions of the proposed model}
    \label{fig:tsne}
\end{figure}

To further investigate and validate our claims of manifold learning, we used t-SNE(\citet{JMLR:v9:vandermaaten08a}) to visualize the bottleneck layer ($\mathcal{C}$) and the linear projection layer ($\mathcal{O}$) (Figure \ref{fig:tsne}) using the training data. We have also color-coded the different points using the RUL labels. 

We can see that adding the autoencoder to reconstruct the input ($\mathcal{X}$) from the bottleneck ($\mathcal{C}$) immediately adds a lot of structure to the representation (Fig \ref{fig:tsne}: B). We hypothesize that this structure that the model is capturing is a domain agnostic manifold in the input space and is helping it generalize across domains. We also note that the smoothness constraint, whose main purpose is to act on the linear projection ($\mathcal{O}$), has also an upstream effect on the learned manifold. $\mathcal{C}$ seems to be more sparse and separated with local clustering of similar colored points.

The linear projection ($\mathcal{O}$) layer is hardly affected by the addition of the autoencoder (Fig \ref{fig:tsne}: D and E). The smoothness constraint doesn't change the overall shape of the manifold, but it rearranges the points in a way that is amenable to final prediction. We can observe a smooth transition from the healthy datapoints on the left to the ones that is at failure (on the right). And we hypothesize that this re-arrangement also helps the model generalize across domains and avoids highly unstable prediction functions.

\section{Conclusion}
In this paper, we develop a novel deep learning architecture (\textit{LAMA-Net}) for Unsupervised Domain Adaptation (UDA) by learning a domain invariant manifold via a combination of losses - regular MSE loss for RUL prediction, MMD loss for aligning the latent representations across source and target domains, reconstruction loss from an autoencoder to enable manifold learning, and a smoothness constraint which enforces uniform continuity in the function that maps the learned manifold to the output space. We then evaluate this architecture on the popular run-to-failure C-MAPSS dataset for turbofan engines and compare against other state-of-the-art alternatives like MMD, CORAL, and DANN. Using RMSE and a asymmetric scoring function, we show that the proposed approach is superior to alternatives in both from the RMSE or Score as well as stability to random seeds perspective. By further analysis into the major architecture decisions, we show how adding an autoencoder enables manifold learning and thereby improves the generalization capability by a large margin. And further the smoothness constraint manage to rearrange the linear projection in a way that makes the final linear layer to predict the RUL with relative ease and stability. We have made available the code at \repo ~ under MIT License.

\medskip

{
\small
\bibliography{ref} 
}

\newpage

\appendix 

\section{Appendix} \label{appendix:details}

\subsection{Datasets - Detailed}
The popular C-MAPSS(\citet{Saxena2008DamagePM}) benchmark dataset of run-to-failure data of turbofan engines, simulated using a thermo-dynamic simulation model. The dataset contains information from 21 sensors as well as 3 operating conditions and is composed of four subsets, namely FD001, FD002, FD003 and FD004, which differ in terms of working conditions, fault modes, life spans, and number of engines (Table \ref{tab: datasets}).

Each subset is further divided into a training set and a testing set. Data sets consists of multiple multivariate time series. Each data set is further divided into training and test subsets. Each time series begins with the engine running normally, and at some point during the series, a defect appears. The fault increases in size in the training set till system failure. The time series in the test set finishes before the system fails. The goal of the competition is to forecast how many operational cycles will remain in the test set before failure, or how many operational cycles will remain in the engine after the final cycle. A vector of accurate Remaining Useful Life (RUL) values for the test data is also provided.

\begin{table*}[!ht]
    \centering
    \caption{Dataset Summary}
    \label{tab: datasets}
    \centering
    \begin{tabular}{llllll}
        \toprule[1.6pt]
        \rule{1pt}{0pt} ~ & FD001 & FD002 & FD003 & FD004\\
        \midrule[1pt]
        Train trajectories & 100 & 260 & 100 & 248\\ 
        Test trajectories & 100 & 259 & 100 & 249  \\ 
        Conditions & 1 (Sea level) & 6 & 1 (Sea level) & 6  \\ 
        Fault Modes (Degradation) & 1 (HPC) & 1 (HPC) & 2 (HPC and Fan) & 2 (HPC and Fan)  \\ 
        Max/min cycles in train set & 362/128 & 378/128 & 25/145 & 543/128 \\ 
        Max/min cycles in test set & 303/31 & 367/21 & 475/38 & 486/19 \\ 
        Train samples & 17731 & 48558 & 21220 & 56815 \\ 
        Test samples & 100 & 259 & 100 & 248\\ 
        \bottomrule[1.6pt]
    \end{tabular}
\end{table*}

\begin{table*}[!ht]
    \centering
    \caption{Selected Hyperparameters}
    \label{tab: hyperparameters}
    \centering
\begin{tabular}{|c|c|c|c|c|}
\hline
\textbf{Hyperparameter}              & \textbf{Ours}                                                                 & \textbf{DANN} & \textbf{MMD} & \textbf{CORAL} \\ \hline
\textit{Window}                      & 40                                                                            & 40            & 40           & 40             \\ \hline
\textit{Epochs}                      & 40                                                                            & 40            & 40           & 40             \\ \hline
\textit{Batch Size}                  & 128                                                                           & 128           & 128          & 128            \\ \hline
\textit{DAST: Attention Dimension}   & 32                                                                            & 32            & 32           & 32             \\ \hline
\textit{DAST: \# of Encoder Layers}  & 3                                                                             & 3             & 3            & 3              \\ \hline
\textit{DAST: \# of Decoder Layers}  & 1                                                                             & 1             & 1            & 1              \\ \hline
\textit{DAST: \# of attention heads} & 4                                                                             & 4             & 4            & 4              \\ \hline
\textit{Squeeze Layer}               & 500-200                                                                       & 500-200       & 500-200      & 500-200        \\ \hline
\textit{Expand Layer}                & 200-500                                                                       & 200-500       & 200-500      & 200-500        \\ \hline
\textit{AutoEncoder}                 & \begin{tabular}[c]{@{}c@{}}GRU(hidden\_size=1, \\ num\_layers=1)\end{tabular} & NA            & NA           & NA             \\ \hline
\textit{Discrepancy Weight} ($\lambda_m$)           & 0.35                                                                          & NA            & 0.2          & 0.2            \\ \hline
\textit{DANN Weight}                 & NA                                                                            & 0.2           & NA           & NA             \\ \hline
\textit{Reconstruction Weight}  ($\lambda_{recon}$)       & 0.2                                                                           & NA            & NA           & NA             \\ \hline
\textit{Smoothness Weight} ($\lambda_{smooth}$)           & 0.35                                                                          & NA            & NA           & NA             \\ \hline
\textit{Local Perturbation}  ($\gamma_{noise}$)          & 0.1                                                                           & NA            & NA           & NA             \\ \hline
\end{tabular} 
\end{table*}

The properties of this dataset regarding the demonstration of prognostic approaches are as follows. 

\begin{itemize}
    \item This is a simulated dataset acquired from commercial software instead of real turbofan engines. Although the degradation processes are simulated as realistic as possible with the consideration of many variability sources, they are still different from real cases.
    \item This dataset contains 21 observations which are composed of different types of features, such as the temperature, pressure, speed, and bleed. Therefore, it is a typical prognostic issue of multi-sensor information fusion. 
    \item This dataset includes six different operational conditions. The time-varying operational conditions cause fluctuation of the degradation trends. Thus, it is suitable for the study of domain adaptation between time-varying operational conditions and degradation trends.
\end{itemize}

\subsection{Data Preprocessing}
The temporal dataset had information coming in from 21 sensors and 3 operational settings. Although we note that 7 sensor values have constant readings in FD001 and FD003, we have still kept all 21 sensor readings and let the model figure out which sensors holds information necessary for prediction.
In order to adapt for different sequence lengths, increase the number of observations the model is trained on, and allow information from past multivariate temporal sequences influence the RUL prediction at any point in time, we employed a sliding window with stride set to one for feature generation. Let the size of the window be $T_w$, then each window in a sequence, $\mathbf{x} = (x_t)^T_{t=0}$, is defined as $\mathcal{W}_t = \left\{ x_{t-T_w}, \dots, x_{t-1} \right\}$. For training the model, the RUL corresponding to the window ($RUL_{t}$) is equal to the RUL at time \textit{t}. No additional feature engineering is done on the input data.

We further normalize the input data by scaling each feature individually such that it is in the (0,1) range using the min-max normalization method
\begin{equation}
norm(x_t^{i,j}) = \frac{x_t^{i,j} - min(x^j)}{max(x^j)-min(x^j)}
\end{equation}
where, $x_t^{i,j}$ is the original i-th datapoint of the j-th input feature at time t and $x^j$ is the vector of all inputs of the j-th input feature.

A typical run-to-failure dataset will have the sensor readings, starting when the machine is in good health, and ending when the machine fails. The RUL labels is something we need to derive from the run-to-failure dataset. For a sequence length of T, a naive approach is to assume a linear decay and set RUL at any time \textit{t}, as remaining time on the run-to-failure data ($R_t = T-t$). But \citet{4711422} has shown that it is a reasonable assumption to estimate RUL as a constant value when the engines operate in good health. Similar to other works (\citet{Hsu}, \citet{s21020418}) in the area using the dataset, we propose to use a piece-wise linear degradation model to define RUL where the initial period of a run-to-failure data is mapped to a constant RUL value ($R_c$) and then switch to a linear degradation based on time.

\begin{equation} 
{RUL}_t = min(T-t, R_c)
\end{equation}

We have chosen $R_c$ to be 125 as suggested by other works using the same dataset. In addition to this we have also scaled the \textit{RUL} to lie in the range (0,1), by dividing the \textit{RUL} with maximum value of \textit{RUL} (which is $R_c$).
\begin{equation}
    scale({RUL}_t) = \frac{{RUL}_t}{R_c} 
\end{equation}

\begin{table*}[!ht]
\centering
\caption{Standard Deviation of Score}
\label{tab: score_sd}
\centering
\begin{tabular}{llccccc}
	\toprule[1.5pt]
	\rule{1pt}{0pt}{Source} & Target & No DA            & DANN             & MMD              & CORAL            & LAMA-Net            \\
	\midrule[1pt]
FD001&FD002&1.31E+19&1.73E+11&1.19E+12&2.09E+17&1.82E+04\\
FD001&FD003&1.48E+05&2.50E+05&3.28E+05&1.28E+06&5.59E+03\\
FD001&FD004&7.93E+21&4.06E+10&2.79E+10&5.88E+14&4.61E+04\\
FD002&FD001&7.67E+09&4.83E+17&1.60E+16&2.08E+16&3.27E+03\\
FD002&FD003&1.55E+08&1.65E+13&2.67E+16&3.24E+18&3.65E+03\\
FD002&FD004&9.53E+28&3.81E+27&7.25E+27&4.42E+26&1.04E+05\\
FD003&FD001&8.40E+03&2.68E+10&1.07E+08&2.07E+09&1.84E+05\\
FD003&FD002&1.11E+19&5.55E+10&3.20E+10&1.21E+18&1.16E+07\\
FD003&FD004&1.13E+20&4.68E+10&8.46E+09&2.73E+17&7.89E+06\\
FD004&FD001&1.58E+10&6.85E+10&8.57E+11&4.69E+14&4.20E+03\\
FD004&FD002&5.95E+23&3.61E+21&2.16E+25&1.74E+30&4.54E+03\\
FD004&FD003&4.84E+09&2.10E+11&1.11E+12&9.67E+12&5.29E+03\\
	\bottomrule[1.5pt]
\end{tabular}
\end{table*}

\subsection{Performance Metrics}
In line with other work using this dataset, the performance of the models was evaluated using the Root Mean Squared Error (RMSE) and an asymmetric scoring function (\citet{Saxena2008DamagePM}). 
\begin{equation} Score = 
\begin{cases}
  \Sigma_{i=1}^{n}{e^{-(E_i/10)} - 1}  & \text{for } $$ E_i<0 $$  ; \\
  \Sigma_{i=1}^{n}{e^{(E_i/13)} - 1} &  \text{for }  $$E_i \geq 0$$ \\
\end{cases}
\end{equation}
where \textit{n} is the number of engines, $E_i = \hat{RUL}_i - RUL_i$ 

\begin{figure}[!ht]
    \centering
    \includegraphics[width=1\linewidth]{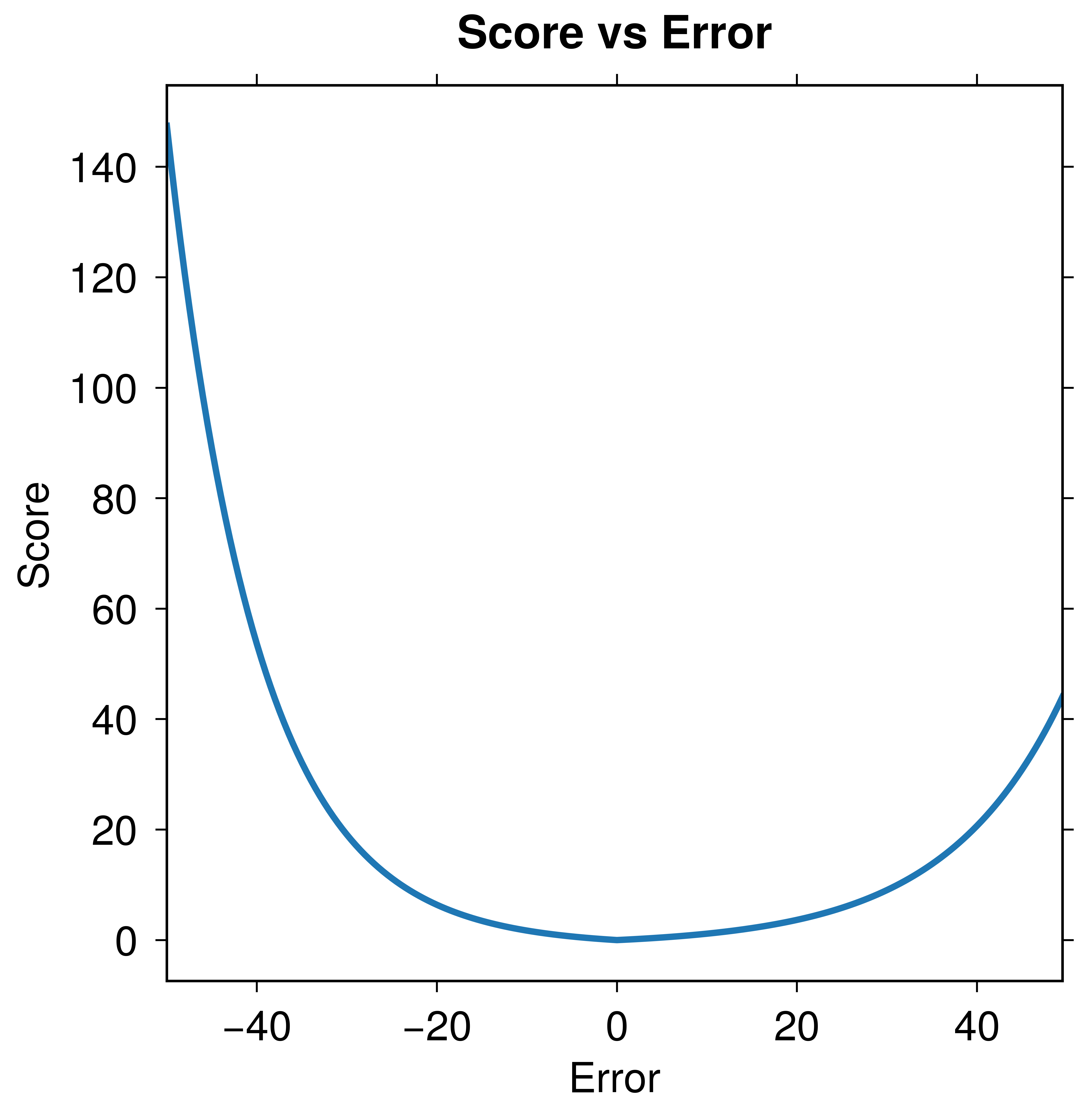}
    \caption{Asymmetric Nature of the Score Metric}
    \label{fig:score}
\end{figure}

Scoring function is structurally an asymmetric function (Figure \ref{fig:score}) that penalises positive errors more than negative errors. RMSE considers an error on either side equally serious, but in RUL prediction, predicting a longer RUL than actual is a much more serious problem than predicting shorter than actual. The scoring function takes care of this nuance in evaluation.
\subsection{Training Procedure}
We choose one of the four C-MAPSS datasets as the source dataset and use our proposed method to learn the RUL on all the other three datasets, without using the RUL labels on those. We use the official training and test split for our validation, and we do create a validation set from the training data using random split with the seed 42. The validation data is only used for monitoring the learning as we do not use any early stopping while training the model. We use mini-batch Gradient Descent for training the model and we randomly select equal number of samples from source and target to form a mini-batch. In all our experiments, we ran the training for 40 epochs with a learning rate decay. We have repeated all experiments using 3 different random seeds (1, 123074, and 2457) and reported the mean and standard deviation to asses the performance as well as stability of the modelling alternatives.

\subsection{Hyperparameter Selection}
For all our experiments, we have chosen the same DAST model with the Squeeze and Expand Layers to maintain consistency in comparison with other techniques. The key hyperparameters for the DAST model are: \textit{hidden size for the attention = 32, number of encoder layers = 3, number of decoder layers = 1} and \textit{number of heads for the attention mechanism = 4}. The \textit{Squeeze Layer} is a multi-layered feed forward network with \textit{500} and \textit{200} units, and the \textit{Expand Layer} is just the opposite with \textit{200} and \textit{500} units such that the output from the \textit{Expand Layer} has the same dimensions as the input to the \textit{Squeeze Layer}. This initial set of \textit{fixed hyperparameters} are fixed such that we have enough capacity in the models to learn the required patterns. Given the structure of the network, we focus our hyperparameter tuning on the core DA mechanism with the below grid:
\begin{itemize}
    \item $\lambda_m, \lambda_r, and \lambda_s$ $\longrightarrow [0.1, 0.2, 0.35, 0.5]$
    \item $\gamma_{noise}$ $\longrightarrow [0.1, 0.01]$
    \item Autoencoder $\longrightarrow ["GRU", "LSTM", "RNN"]$
\end{itemize}

Instead of selecting a configuration for each source-target combinations, we chose a set of hyperparameters which performed better on the validation source RUL across all the source-target combinations. This helps us come up with a set of hyperparameters which has more generalization capabilities. 
\subsection{Final Hyperparameters}
The final set of hyperparameters used in the experiments are in Table \ref{tab: hyperparameters}.

\subsection{Additional Results}
The standard deviation across three runs with different random seeds for the Score is in Table \ref{tab: score_sd}.

\end{document}